\documentclass[runningheads]{llncs}
\usepackage{graphicx}
\usepackage{amsfonts} 
\usepackage{amssymb}
\usepackage{array} 
\usepackage{pict2e}
\usepackage{tikzscale}
\usepackage{bbm}
\usepackage{amsmath}
\usepackage{subcaption}
\usepackage{tikz}
\usepackage{hyperref}
\usepackage{multirow}
\usepackage{pdfpages}
\begin{document}
\title{DRIM: Learning Disentangled Representations from Incomplete Multimodal Healthcare Data}
\titlerunning{Disentangled Representations from Incomplete Multimodal Healthcare Data}

\author{Lucas Robinet\inst{1, 2, 3}\and Ahmad Berjaoui\inst{2}\and Ziad Kheil\inst{1,3} \and Elizabeth Cohen-Jonathan Moyal\inst{1, 3}}
\authorrunning{Robinet et al.}
\institute{IUCT-Oncopole Institut Claudius Regaud, Toulouse, France \\ \email{\{robinet.lucas,moyal.elisabeth\}@iuct-oncopole.fr}\\ \and  IRT Saint Exupéry, Toulouse, France \\ \email{ahmad.berjaoui@irt-saintexupery.com} \\ \and INSERM UMR 1037, Cancer Research Center of Toulouse (CRCT), Toulouse, France \\ \email{ziad.kheil@inserm.fr} }

\maketitle              
\begin{abstract}
Real-life medical data is often multimodal and incomplete, fueling the growing need for advanced deep learning models capable of integrating them efficiently. 
The use of diverse modalities, including histopathology slides, MRI, and genetic data, offers unprecedented opportunities to improve prognosis prediction and to unveil new treatment pathways. 
Contrastive learning, widely used for deriving representations from paired data in multimodal tasks, assumes that different views contain the same task-relevant information and leverages only shared information. 
This assumption becomes restrictive when handling medical data since each modality also harbors specific knowledge relevant to downstream tasks.
We introduce DRIM, a new multimodal method for capturing these shared and unique representations, despite data sparsity. 
More specifically, given a set of modalities, we aim to encode a representation for each one that can be divided into two components: one encapsulating patient-related information common across modalities and the other, encapsulating modality-specific details. 
This is achieved by increasing the shared information among different patient modalities while minimizing the overlap between shared and unique components within each modality.
Our method outperforms state-of-the-art algorithms on glioma patients survival prediction tasks, while being robust to missing modalities. To promote reproducibility, the code is made publicly available at \href{https://github.com/Lucas-rbnt/DRIM}{https://github.com/Lucas-rbnt/DRIM}
\end{abstract}
\section{Introduction}
\label{sec:intro}
Cancer diagnosis and prognosis are usually determined by an ensemble of data obtained from various sources including MRI, histopathology slides, and molecular profiles.
Combining these modalities allows to derive a panoply of biomarkers, offering a deeper understanding of the tumor landscape. 
Thereby, it enables clinicians to provide patients with the utmost support throughout their pathology \cite{gallego_nonsurgical_2015}.
In fields beyond medicine, there are several deep learning models and techniques capable of merging multimodal data \cite{girdhar_imagebind_2023,oord_representation_2019,zadeh_tensor_2017,han_improving_2021}.
However, in a medical context, the prevalent challenge of integrating incomplete heterogeneous modalities steers deep learning techniques towards simpler multimodal or even unimodal-based models \cite{wetstein_deep_2022,bae_radiomic_2018}.
While some multimodal approaches rely on late fusion \cite{steyaert_multimodal_2023} i.e. averaging outputs from each modality, most popular methods focus on merging modalities in the latent space \cite{vale-silva_long-term_2021,cheerla_deep_2019,braman_deep_2021,chen_pathomic_2020,huang_salmon_2019}. 
Indeed, MultiSurv \cite{vale-silva_long-term_2021} explores the utility of straightforward fusion methods such as mean, sum, product, and maximum, testing their capacity to handle missing modalities within survival prediction tasks. 
Although simple, another fusion technique that has showcased its effectiveness in the medical domain involves the concatenation of the representations \cite{steyaert_multimodal_2023}.
Finally, \cite{zadeh_tensor_2017} introduced a parameterized approach called Tensor, which leverages the outer product of representations. 
This method was later applied in \cite{chen_pathomic_2020} to combine histology slides and genomic data.
Nonetheless, this method falls short in naturally accommodating missing modalities and faces a critical limitation as the amount of parameters it requires increases exponentially with each added modality.
This excessively high number of parameters hinders its effectiveness for combining multiple modalities as it drastically limits the representation dimensions.

Beyond mere fusion, leveraging auxiliary losses on representations to induce specific behaviors is a strategy adopted in previous works \cite{cheerla_deep_2019,braman_deep_2021,zhou_attentive_2023}. 
Hence, \cite{cheerla_deep_2019,zhou_attentive_2023} embed a correlation function within representations to promote consistency. 
Conversely, \cite{braman_deep_2021} adopts a divergent strategy, denoted multimodal orthogonalization (MMO), striving for their representations to minimize redundancy and overlapping information across modalities.

In this paper, unlike to prior work, we make the natural assumption that information contained in a modality splits into patient-related aspects shared across modalities and unique features specific to each modality.
Our key contributions are: (i) employing a pair of encoders for each modality —one shared, one unique. 
Optimizing the shared encoder can be viewed through the lens of Information Bottleneck \cite{tishby_deep_2015}, ensuring that representations are rich in patient-specific information yet minimally captures the modality's characteristics. 
Concurrently, unique encoders aim to reduce mutual information between their representation and the corresponding shared one. 
(ii) Leveraging these disentangled representations, we introduce a dual-scale fusion approach: it initiates with the aggregation of shared information, later combined with unique information to craft a comprehensive representation pertinent for survival analysis.
(iii) We employ an attention-based, scalable and parameterized fusion that intuitively models interactions between representations and natively manages missing modalities.

Gliomas, with their diverse prognosis biomarkers, range of grades, and complexity as described in the WHO 2021 classification \cite{louis_2021_2021}, present a compelling use case for showcasing DRIM's ability to effectively capture and disentangle intricate information.

\section{Method}
\label{sec:method}
\begin{figure}[!ht]
    \centering
    \begin{subfigure}[b]{0.5\textwidth}
        \centering
        \includegraphics[width=\textwidth, height=7.37cm]{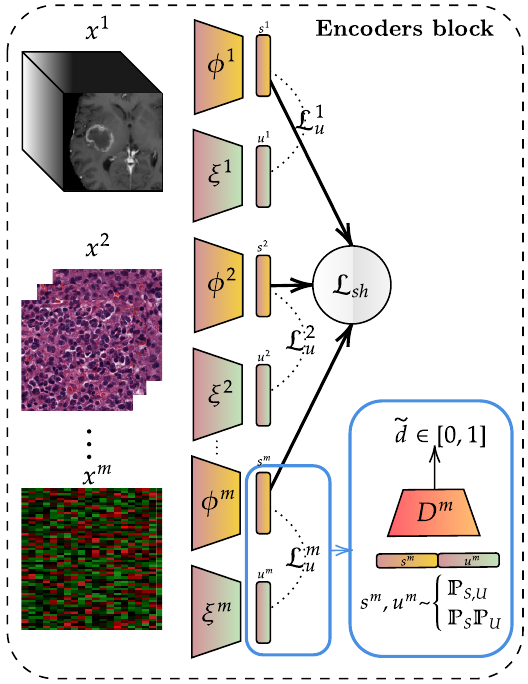}
        \caption{Encoder training (Sec~\ref{subsec:shared},~\ref{subsec:unique})}
        \label{fig:encodersblock}
    \end{subfigure}%
    \begin{subfigure}[b]{0.5\textwidth}
        \begin{subfigure}[b]{\textwidth}
            \centering
            \includegraphics[width=0.9\textwidth, height=3.35cm]{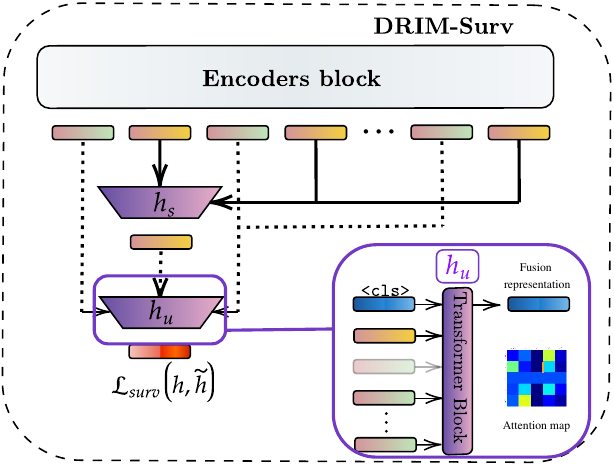}
            \caption{Survival DRIM-Surv (Sec.~\ref{subsec:task})}
            \label{fig:drimsurv}
        \end{subfigure}
        \begin{subfigure}[b]{\textwidth}
            \centering
            \includegraphics[width=0.9\textwidth, height=3.35cm]{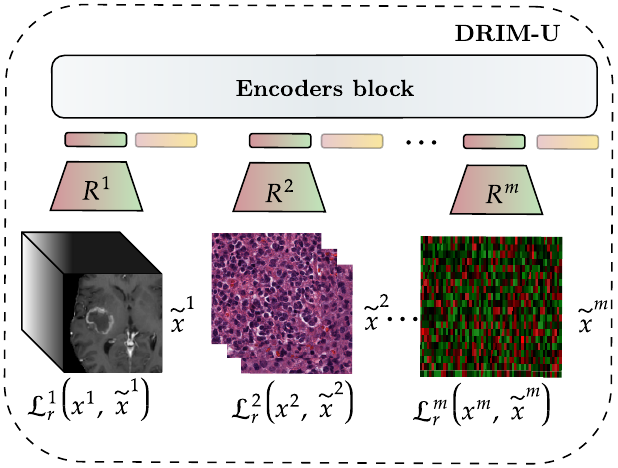}
            \caption{Unsupervised DRIM-U (Sec.~\ref{subsec:task})}
            \label{fig:drimu}
        \end{subfigure}
    \end{subfigure}
\caption{An overview of the proposed method. (a) describes the learning of shared and unique components, while (b) and (c) propose a supervised and an unsupervised alternative to couple these representations to a specific task.}
\label{fig:framework}
\end{figure}
\noindent
Consider a training minibatch of $N$ samples and $M$ modalities, with $x^m = \{x_i^m\}_{i=1}^{N}$ representing data from the $N$ samples for a specific modality $m$.
For each modality $m$, let $\phi^m: x^m \mapsto s^m \in \mathcal{R}^{N\times d}$ and $\xi^m: x^m \mapsto u^m \in \mathcal{R}^{N\times d}$ denote the shared and unique encoders respectively, with $s_i^m$ and $u_i^m$, being the representations of the $m$-th modality for the $i$-th sample in the minibatch. 
DRIM makes it possible to learn disentangled representations while being flexible to a range of tasks. 
Indeed, the DRIM-Surv approach (Fig.~\ref{fig:drimsurv}) operates by fusing representations at two distinct scales to perform supervised end-to-end survival training. 
Alternatively, DRIM-U provides a wholly unsupervised option for learning meaningful shared and unique representations as illustrated in Fig.~\ref{fig:drimu}.
In both scenarios, the DRIM loss can be dissected into three distinct components as illustrated in Eq.~\ref{eq:drim}: a task-related term, a shared term, and a unique term.
\begin{align}
\label{eq:drim}
\mathcal{L}_{DRIM} = \mathcal{L}_{T} + \mathcal{L}_{sh} + \gamma \mathcal{L}_u
\end{align}
\subsection{Shared Loss}
\label{subsec:shared}

To maximize mutual information between shared representations $s^m$, traditional methods optimize the InfoNCE objective \cite{oord_representation_2019} for each modality pair \cite{girdhar_imagebind_2023,cheerla_deep_2019}, leading to a computational complexity of $O(M^2)$.
Thereby, in Eq. \ref{eq:shared}, we propose a new well-suited shared cost function (see Fig.~\ref{fig:encodersblock}).
Inspired by supervised contrastive learning \cite{khosla_supervised_2020} wherein class labels facilitate generalization to multiple positive pairs, we stipulate that each representation $s^m$ finds positive pairs among representations stemming from distinct modalities yet originating from the same patient.
Thus, all shared representations are concatenated to create the matrix $S =[s_1^1, s_2^1, \cdots s_1^2, s_2^2, \cdots s_{N-1}^M, s_N^M]^T \in \mathcal{R}^{N\cdot M \times d}$.
Let $J := \{1, ..., N \cdot M\}$ represent the indices of representations, $B(j) = \{b \in J: b\neq j, \kappa(b) = \kappa(j)\}$ define the set of its positive pairs, where $\kappa(j)$ maps $j$ to the patient it belongs to, $\tau \in \mathcal{R}^+$ is a scalar temperature parameter. The shared loss term is formulated as follows:
\begin{align}
\label{eq:shared}
\mathcal{L}_{sh} = \frac{1}{N\cdot M} \sum_{j \in J} \frac{-1}{M - 1} \sum_{b \in B(j)} \log \frac{exp(s_j^T s_b /\tau)}{\sum_{\substack{k \in J \\ k\neq j}} exp (s_j^T s_k / \tau)} 
\end{align}
\subsection{Unique Loss}
\label{subsec:unique}

Until now, we have successfully derived shared representations $\{s^m\}_{m=1}^M$.
The unique representation $u^m$, extracted by $\xi^m$, parameterized by $\psi^m$, should highlight features specific to its modality $m$ without encompassing the information contained in $s^m$. 
To this end, we use an adversarial objective \cite{goodfellow_generative_2014}, akin to the approach outlined in \cite{sanchez_learning_2020}, to minimize the mutual information $I(s^m,u^m)$. 
A discriminator $D^m$, parameterized by $\theta_m$ is trained for each modality $m$ to differentiate representations sampled from the joint distribution from those sampled from the product of the marginals as shown in Fig.~\ref{fig:encodersblock}. 
Simulating samples from the product of the marginals $\mathbb{P}_{S}\mathbb{P}_{U}$ involves shuffling the batch of shared representations before pairing with unique ones as in \cite{brakel_learning_2017}. 
This adversarial objective outlined in Eq. \ref{eq:unique} drives the minimization of the Jensen-Shannon divergence $D_{JS}(\mathbb{P}_{S,U}\|\mathbb{P}_{S}\mathbb{P}_{U}))$, as evidenced in \cite{goodfellow_generative_2014}.
\begin{align}
\mathcal{L}_u(\theta, \psi) = - &\sum_{m=1}^{M} \Big[\mathbb{E}_{\mathbb{P}_{S,U}}[\log(1 - D^m(s^m,u^m));\theta^m]\notag \\ & + \mathbb{E}_{\mathbb{P}_{S}\mathbb{P}_{U}}[\log D^m(s^m,u^m);\theta^m] + \mathbb{E}_{\mathbb{P}_{S,U}}[\log D^m(s^m,u^m);\psi^m]\Big]
\label{eq:unique}
\end{align}
\subsection{Task Loss}
\label{subsec:task}
Merely separating shared and unique components does not guarantee the relevance of the learned unique representations. 
The training of $\xi^m$ might still collapse, risking a meaningless latent space.
Hence, to ensure we capture of pertinent information, each unique encoder is tied to a specific task.
Indeed, in DRIM-Surv, unique encoders are also linked to a survival task, compelling them to identify modality-specific relevant features.
To delve deeper, let $\{x_i, t_i, \delta_i\}_{i=1}^{N}$ denote our minibatch where $t_i$ represents the time of last follow-up and $\delta_i$ indicates if the event was observed ($\delta_i = 1$) or censored ($\delta_i = 0$). Time is divided into $P$ equidistant fixed intervals, and a neural network $f(x) = [\tilde{h}_1, \cdots, \tilde{h}_P] \in [0,1]^P$ aims to estimate the conditional hazard probability for each time interval. Let also $\upsilon: \mathcal{R} \rightarrow \{1, \cdots, P\}$ be a function that maps a follow-up $t$ to the the interval in which it lies. The survival loss function used is an extension of \cite{gensheimer_scalable_2019} provided by \cite{kvamme_continuous_2021} allowing predicted hazards to be non-proportional.
\begin{align}
\mathcal{L}_{T} = - \frac{1}{N}\sum_{i=1}^N \sum_{k=1}^{\upsilon(t_i)} \mathbbm{1}_{\{\upsilon(t_i) = k, \delta_i = 1\}} \log{\tilde{h}_k} + (1 - \mathbbm{1}_{\{\upsilon(t_i) = k, \delta_i = 1\}}) \log{(1 - \tilde{h}_k)}
\end{align}
Moreover, with the DRIM-U alternative, each $\xi^m$ is also optimized to encode and decode inputs through a specific decoder $R^m(\xi^m(x^m)) = \tilde{x}^m$, minimizing reconstruction error $\mathcal{L}_{T} = \sum_{m=1}^M \mathcal{L}_r^m$ and $\mathcal{L}_r^m = \sum_{i} \|\tilde{x_i}^m - x_i^m\|_2^2$ as in Fig.~\ref{fig:drimu}.

\subsection{Multimodal Fusion}
\label{subsec:fusion} 
In DRIM-Surv, we merge representations on two scales: first, $h_s$ fuses shared representations into a global shared vector $s \in \mathcal{R}^{N \times d}$; then, this is combined with unique representations via $h_u$ to form a unified patient embedding, depicted in Fig.~\ref{fig:drimsurv}.
In both fusion, a masked transformer processes each representation as a sequence token, with a classification token \texttt{<cls>} added at the beginning, akin to a ViT approach \cite{dosovitskiy_image_2020}.
The output of the transformer for this token acts as the fused representation, as shown in Fig.~\ref{fig:drimsurv}.
Specifically, $h_u$ aggregates the $M$ unique representations plus the global shared one $s$ through $h_u([u^1, \cdots, u^M, s]) = softmax\big(\frac{Q_{cls}K^T}{\sqrt{D}}\big)V$
where $Q_{cls} = q_{cls}W_Q$, $K = CW_K$ and $V = CW_V$ are linear transformations of the learnable query token \texttt{<cls>}, and representations $C=[\texttt{<cls>}, u^1,\cdots, u^M, s]$ respectively and $D$ the projection dimension.
In practice, we use multiple attention heads.
This fusion, termed MAFusion, offers scalability, computational efficiency, robust handling of missing modalities through masking and provides interpretability via attention maps.
\section{Experiments}
\label{sec:expe}
This study leverages multimodal glioma data from TCGA \cite{chang_cancer_2013}, specifically GBM and LGG collections, across four modalities: DNA methylation, (DNAm), RNA sequencing (RNA), histopathological slides (WSI) and MRI. 
MRI data was completed from the BraTS dataset \cite{baid_rsna-asnr-miccai_2021}, selecting only TCGA-listed patients. 
Focusing on patient with at least two modalities, we kept 881 samples, finding coverage rates of 95\% for DNAm, 89\% for WSI, 75\% for RNA, and 18\% for MRI.
\subsubsection{Genomics (DNAm and RNA):}
DNA methylation data were processed into 25,978 Beta values using the same method as in \cite{vale-silva_long-term_2021}.
RNA data were processed to retain protein-coding genes with variance over 0.1, resulting in 16,304 FPKM-UQ normalized genes, then log-transformed and z-score normalized.
Both genomics encoders comprise six 1D convolutional layers with channels increasing from 1 to 256. 
Each layer applies GELU activation \cite{hendrycks2023gaussian}, batch normalization, and dropout, ending with Adaptive Average Pooling to produce a fixed-size embedding. 
\subsubsection{WSI:} Slides were subsampled, and a mask designed to isolate tissue zones was extracted using an OTSU filter.
From these, 1000 high-resolution patches of size 256x256 were randomly selected, of which 100 were chosen based on a custom HSV intensity score (Fig. S1). A SimCLR technique \cite{chen_simple_2020} was employed to train a ResNet34 \cite{He_2016_CVPR} on the chosen patches.
Then, the WSI encoder, a single-layer transformer with 8 attention heads and no positional encoding, projects 10 randomly selected patch representations into a meaningful embedding.
\subsubsection{MRI scans:} Segmentation masks were used to crop a $64\times 64\times 64$ voxel region surrounding the tumor in Gd-T1w and FLAIR MRI scans.
A SimCLR approach \cite{chen_simple_2020}, similar to the WSI one, is used to train a MONAI ResNet10-3D \cite{cardoso_monai_2022} across the BraTS dataset \cite{baid_rsna-asnr-miccai_2021}. 
In our DRIM experiments, we froze the encoder weights and added a 256-unit linear layer with dropout and GELU activation \cite{hendrycks2023gaussian}, followed by a dense layer.
\subsubsection{Implementation details:}
Models were trained using an AdamW(lr=$1e^{-3}$, wd=$1e^{-2}$) optimizer over 30 epochs with a Cosine Annealing Scheduler and a batch size of $24$. 
For all experiments, the representation size was set to 128, except in those involving Tensor \cite{zadeh_tensor_2017}, where it was reduced to 32 due to memory constraints—approaching 34,500 million parameters at a dimension of 128 (Table S1).
The final layer of each model consisted of 20 units with sigmoid activation.
The MAFusion utilized 16 attention heads of size 64 and a dense layer of size 128.
In DRIM experiments, discriminators were also optimized via AdamW (lr=$1e^{-3}$, wd=$3e^{-4}$), and the loss coefficient $\gamma$ is set empirically to $0.8$.
For the DRIM-U evaluation, encoders were pre-trained for 30 epochs as detailed in Section \ref{sec:method} and then frozen. 
Subsequently, a DRIM-Surv module was fine-tuned on top of these representations over 10 epochs for survival prediction.
\subsubsection{Evaluation:}
The dataset is divided into a training set and a test set, with the latter comprising 20\% of the entire dataset. 
We use five-fold cross-validation on the training set, with performance metrics reported as mean ± standard deviation on the test set from models trained on 4 (out of 5) training folds. 
Evaluation metrics include the Concordance Index (C-index) \cite{antolini_time-dependent_2005}, Integrated Brier Score (IBS) \cite{gerds_consistent_2006,graf_assessment_1999}, and Integrated Negative Binomial Log-likelihood (INBLL) \cite{graf_assessment_1999,kvamme_continuous_2021}.
\section{Results and discussions}
\label{sec:results}
\subsection{Survival performance}
\begin{table}[!ht]
\centering
\caption{Comparison of mean performance metrics among survival models.}
\label{tab:results}
\begin{tabular}{|c||l||ccc|}
\hline
Method                    & Fusion       & C-index ($\uparrow$) & Integrated BS ($\downarrow$) & INBLL ($\downarrow$) \\ \hline \hline
\multirow{7}{*}{Vanilla}  & Mean \cite{vale-silva_long-term_2021}   &   $0.763\pm0.010$     &     $0.110\pm0.005$                   &   $0.352\pm0.026$    \\
                          & MMean \cite{cheerla_deep_2019} &    $0.757\pm0.008$      &      $0.110\pm0.006$                  &   $0.361\pm0.030$    \\
                          & Conc. \cite{steyaert_multimodal_2023} &  $0.760\pm0.006$       &          $0.095\pm0.005$              & $0.319\pm0.019$      \\
                          & Sum \cite{vale-silva_long-term_2021}  &    $0.755\pm0.007 $     &               $0.097\pm0.004$          &   $0.328\pm0.030$ \\
                          & Max \cite{vale-silva_long-term_2021}   &   \underline{0.766 $\pm$ 0.006}       &      $0.097\pm0.006$                   &   $0.313\pm0.026$    \\
                          & Tensor \cite{chen_pathomic_2020} &  $0.765\pm0.006$       &  $0.093\pm0.005$                   &  $0.292\pm0.015$     \\
                          & MAFusion     &   $0.728\pm0.009$      &  $0.100\pm0.005$                      &    $0.382\pm0.029$   \\ \hline \hline
\multirow{4}{*}{MMO Loss \cite{braman_deep_2021}} & Mean \cite{vale-silva_long-term_2021}   &    $ 0.759\pm0.008$     &           $0.111\pm0.005$             &  $0.358\pm0.029$     \\
                          & Conc. \cite{steyaert_multimodal_2023} &  0.764 $\pm$ 0.007       &        0.093 $\pm$ 0.004                &   0.311 $\pm$ 0.018    \\
                          & Tensor \cite{chen_pathomic_2020} &   0.764 $\pm$ 0.004      &   0.090 $\pm$ 0.007                     &   0.286 $\pm$ 0.026     \\
                          & MAFusion     &   0.728 $\pm$ 0.011      &         0.100 $\pm$ 0.008               &    0.389 $\pm$ 0.052   \\ \hline \hline
\multirow{1}{*}{Sh. Loss only}  
                          & MAFusion &   0.750 $\pm$ 0.016      &          \underline{0.090 $\pm$ 0.006}              &    0.324 $\pm$ 0.022   \\ \hline \hline
DRIM-U (ours)             & MAFusion     &    0.763 $\pm$ 0.006     &      0.091 $\pm$ 0.06                  &  \underline{0.286 $\pm$ 0.021}     \\
DRIM-Surv (ours)          & MAFusion     &  \textbf{0.774$\pm$0.006}       &        \textbf{0.086$\pm$0.004}                &       \textbf{0.285$\pm$0.016} \\ \hline 
\end{tabular}
\end{table}

The data in Table~\ref{tab:results} demonstrate that the DRIM-Surv approach outperforms several other methods across all three assessed metrics. 
This superiority highlights the strong ability of our method to accurately discriminate patient outcomes (C-index) as well as to reliably forecast survival probabilities over time (IBS, INBLL).
Notably, DRIM-Surv not only slightly exceeds the performance of Tensor \cite{zadeh_tensor_2017} in all metrics, but it also offers advantages in terms of scalability to additional modalities, reduced training time, and a significantly lower parameter count (Table S1 and Table S2).
Indeed, except for the new encoders parameters, incorporating an additional modality into the DRIM framework leaves its parameter count unchanged, while the parameter total for Tensor \cite{zadeh_tensor_2017} would escalate from 38 million—already near 70 times larger than those of our MAFusion—to exceed one billion (Table S1).
The sensitivity analysis on the $\gamma$ coefficient illustrates that, beyond the neural network architecture, our method effectively learns to disentangle the representations of each modality, thereby enhancing the reliability of the final prognosis (Table S3).
Finally, our unsupervised DRIM-U approach also gives convincing results simply by being fine-tuned on a few epochs, validating the flexibility of our method. 
\subsection{Stratifying high-risk vs low-risk patients}
Additionally, we evaluate the proficiency of our model to stratify patients into high-risk and low-risk categories, a crucial step towards advancing personalized medicine.
Concretely, for a given model, patients are divided into two groups based on their cumulative predicted hazards: those in the top half are classified as high-risk, while those in the bottom half are considered low-risk.
Survival curves for each group are derived using Kaplan-Meier estimators, with a log-rank test evaluating their differences and stratification efficacy.
For DRIM-Surv, The logrank test yields a p-value of $2.9e^{-23}$, against $4.2e^{-22}$ for the Tensor \cite{zadeh_tensor_2017} w/ MMO \cite{braman_deep_2021} method, emphasizing the ability of our model to distinguish survival outcomes between the two risk groups (Fig. S2).
\subsection{Robustness across varied input modality combinations}
One of the primary benefits of our approach lies in its ability to smoothly manage missing modalities during inference, avoiding the use of zero-filled tensors required by other methods \cite{zadeh_tensor_2017,braman_deep_2021,steyaert_multimodal_2023}.
Indeed, DRIM goes beyond simple arranged modality combination; it adeptly models their interactions and efficiently extracts crucial information, showcasing advanced data understanding.
\begin{table}[!ht]
\centering
\caption{Analysis of CS ($\frac{C^{index} + (1 - IBS)}{2}$) metrics \cite{steyaert_multimodal_2023} on test samples containing at least all specified modalities, setting any extra modalities to zero for inference. The table displays only the top three competing models. For each combination of inputs, the percentage of occurrence in the training dataset is given (\% train).}
\label{tab:robust}
\begin{tabular}{|l||c||c||c||c|}
\hline
Modalities (\% train)     & Max               & \begin{tabular}[c]{@{}c@{}}Conc.\\ w/ MMO\end{tabular}         & \begin{tabular}[c]{@{}c@{}}Tensor\\ w/ MMO\end{tabular}         & \textbf{\begin{tabular}[c]{@{}c@{}}DRIM-Surv\\ (ours)\end{tabular}} \\ \hline \hline
WSI,MRI    (2\%)   &     \underline{0.702 $\pm$ 0.019}              &     0.643 $\pm$ 0.039              &      0.619 $\pm$ 0.060             & \textbf{0.796$\pm$0.016}                                                  \\
WSI,RNA   (2\%)    &          0.795 $\pm$ 0.011         &       \underline{0.816 $\pm$ 0.010}            &    0.811 $\pm$ 0.021               & \textbf{0.829$\pm$0.006}                                                   \\
DNAm,WSI  (17\%)    &         0.798 $\pm$ 0.004          &      \underline{0.810 $\pm$ 0.007}             &     $0.804\pm0.010$              & \textbf{0.832$\pm$0.008}                                                    \\
DNAm,RNA    (9\%)  &         0.843 $\pm$ 0.006            &      0.848 $\pm$ 0.009             &      \textbf{0.860$\pm$0.009}               & \underline{0.858 $\pm$ 0.011}                                                  \\
DNAm,MRI   (2\%)   &            0.740 $\pm$ 0.018         &       \underline{0.740 $\pm$ 0.014}            &       0.740 $\pm$ 0.015                & \textbf{0.794$\pm$0.025}                                                   \\
RNA,MRI (0\%)      &        0.783 $\pm$ 0.039           &        \underline{0.803 $\pm$ 0.041}           &     0.755 $\pm$ 0.033              & \textbf{0.874$\pm$0.024}                                                   \\ \hline \hline
DNAm,WSI,RNA(55\%) &       0.838 $\pm$ 0.007             &           0.839 $\pm$ 0.010        &       \textbf{0.853$\pm$0.009}            & \underline{0.850 $\pm$ 0.015}                                                   \\
DNAm,WSI,MRI (3\%)&      0.767 $\pm$ 0.031             &      \underline{0.779 $\pm$ 0.022}             &    0.761 $\pm$ 0.029               & \textbf{0.838$\pm$0.020}                                                     \\
DNAm,RNA,MRI(1\%)&       0.824 $\pm$ 0.033            &     \underline{0.870 $\pm$ 0.036}              &     0.823 $\pm$ 0.047              & \textbf{0.874$\pm$0.032}                                                   \\
WSI,RNA,MRI (1\%) & 0.828 $\pm$ 0.019 & \underline{0.830 $\pm$ 0.027} & 0.771 $\pm$ 0.043 & \textbf{0.892$\pm$0.026}                                                   \\ \hline \hline
All 4 modalities (8\%)& \underline{0.886 $\pm$ 0.014} & 0.885 $\pm$ 0.017 & 0.828 $\pm$ 0.050 & \textbf{0.904$\pm$0.029}\\ \hline
\end{tabular}
\end{table}
Table~\ref{tab:robust} reveals that while other models show comparable performance on their most frequent training combination (55\%), DRIM uniquely excels across all input combinations, surpassing its competitors. For instance, when inferring on WSI and MRI data, DRIM attains a CS score of $0.796 \pm 0.016$, far surpassing the $0.619 \pm 0.060$ achieved by Tensor \cite{zadeh_tensor_2017} w/ MMO \cite{braman_deep_2021}.
Furthermore, our method is the only one that almost systematically improves its performance when a modality is added to a given combination.
This distinction demonstrates that, whereas other models rely heavily on \textit{learning} from a fixed and ordered sequence of modalities, DRIM-Surv stands out in its capability to specifically extract and combine information from diverse inputs.
\section{Conclusions and perspectives}
\label{sec:conclu}
We propose DRIM, an approach for disentangling data representations from highly heterogeneous and incomplete sources.
Our method not only enhances the forecasting of patient outcomes through an adept handling of input modalities but also demonstrates improved scalability and flexibility with the addition of new modalities, unlike classic multimodal techniques.
This versatility ensures its applicability to a broad spectrum of datasets featuring diverse modalities and tasks.
Finally, we aspire that the ongoing efforts to disentangle learned representations of modalities will pave the way for deeper analysis and enhanced interpretability.
This development promises to unlock precise, individualized patient assessments, thereby bridging the gap to personalized medecine and practical clinical adoption.
\bibliographystyle{splncs04}
\bibliography{bibliography}
\includepdf[pages=-]{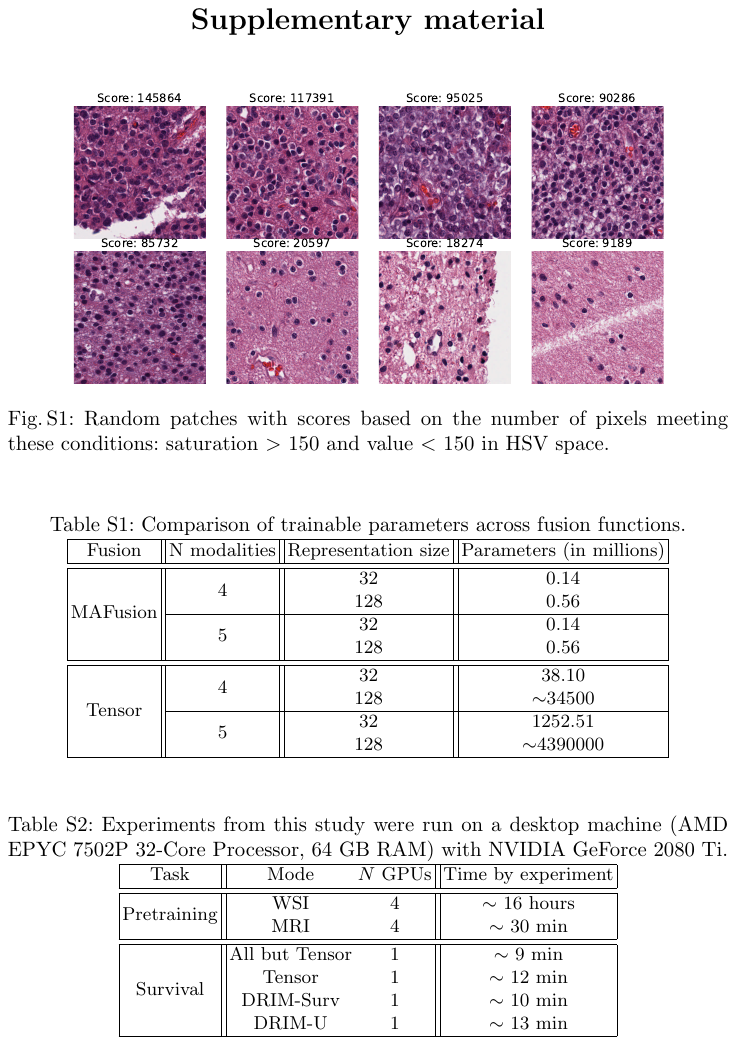}
\end{document}